\def\year{2019}\relax
\documentclass[letterpaper]{article} 
\usepackage{aaai19}  
\usepackage{times}  
\usepackage{helvet}  
\usepackage{courier}  
\usepackage{url}  
\usepackage{graphicx}  
\usepackage{amsmath,amssymb,bm}
\usepackage{algorithm}
\usepackage{algorithmic}
\graphicspath{{figures/}}
\newcommand{\citet}[1]{\citeauthor{#1}~\shortcite{#1}}
\newcommand{\citep}{\cite}
\newcommand{\citealp}[1]{\citeauthor{#1}~\citeyear{#1}}

\def\x{\mathbf{x}}
\def\a{\mathbf{a}}

\def\z{\mathbf{z}}
\def\w{\mathbf{w}}

\DeclareMathOperator{\ab}{\mathbf{a}}
\DeclareMathOperator{\bb}{\mathbf{b}}
\DeclareMathOperator{\cbb}{\mathbf{c}}
\DeclareMathOperator{\db}{\mathbf{d}}
\DeclareMathOperator{\eb}{\mathbf{e}}
\DeclareMathOperator{\fb}{\mathbf{f}}
\DeclareMathOperator{\gb}{\mathbf{g}}
\DeclareMathOperator{\hb}{\mathbf{h}}
\DeclareMathOperator{\ib}{\mathbf{i}}
\DeclareMathOperator{\jb}{\mathbf{j}}
\DeclareMathOperator{\kb}{\mathbf{k}}
\DeclareMathOperator{\lb}{\mathbf{l}}
\DeclareMathOperator{\mb}{\mathbf{m}}
\DeclareMathOperator{\nbb}{\mathbf{n}}
\DeclareMathOperator{\ob}{\mathbf{o}}
\DeclareMathOperator{\pb}{\mathbf{p}}
\DeclareMathOperator{\qb}{\mathbf{q}}
\DeclareMathOperator{\rb}{\mathbf{r}}
\DeclareMathOperator{\sbb}{\mathbf{s}}
\DeclareMathOperator{\tb}{\mathbf{t}}
\DeclareMathOperator{\ub}{\mathbf{u}}
\DeclareMathOperator{\vb}{\mathbf{v}}
\DeclareMathOperator{\wb}{\mathbf{w}}
\DeclareMathOperator{\xb}{\mathbf{x}}
\DeclareMathOperator{\yb}{\mathbf{y}}
\DeclareMathOperator{\zb}{\mathbf{z}}

\DeclareMathOperator{\Ab}{\mathbf{A}}
\DeclareMathOperator{\Bb}{\mathbf{B}}
\DeclareMathOperator{\Cb}{\mathbf{C}}
\DeclareMathOperator{\Db}{\mathbf{D}}
\DeclareMathOperator{\Eb}{\mathbf{E}}
\DeclareMathOperator{\Fb}{\mathbf{F}}
\DeclareMathOperator{\Gb}{\mathbf{G}}
\DeclareMathOperator{\Hb}{\mathbf{H}}
\DeclareMathOperator{\Ib}{\mathbf{I}}
\DeclareMathOperator{\Jb}{\mathbf{J}}
\DeclareMathOperator{\Kb}{\mathbf{K}}
\DeclareMathOperator{\Lb}{\mathbf{L}}
\DeclareMathOperator{\Mb}{\mathbf{M}}
\DeclareMathOperator{\Nb}{\mathbf{N}}
\DeclareMathOperator{\Ob}{\mathbf{O}}
\DeclareMathOperator{\Pb}{\mathbf{P}}
\DeclareMathOperator{\Qb}{\mathbf{Q}}
\DeclareMathOperator{\Rb}{\mathbf{R}}

\DeclareMathOperator{\Tb}{\mathbf{T}}
\DeclareMathOperator{\Ub}{\mathbf{U}}
\DeclareMathOperator{\Vb}{\mathbf{V}}
\DeclareMathOperator{\Wb}{\mathbf{W}}
\DeclareMathOperator{\Xb}{\mathbf{X}}
\DeclareMathOperator{\Yb}{\mathbf{Y}}
\DeclareMathOperator{\Zb}{\mathbf{Z}}

\newcommand{\thetab}{{\bm{\theta}}}

\ifx\proof\undefined
\newenvironment{proof}{\par\noindent{\bf Proof\ }}{\hfill\BlackBox\\[2mm]}
\fi

\ifx\theorem\undefined
\newtheorem{theorem}{Theorem}
\fi
\ifx\example\undefined
\newtheorem{example}{Example}
\fi
\ifx\lemma\undefined
\newtheorem{lemma}[theorem]{Lemma}
\fi

\ifx\corollary\undefined
\newtheorem{corollary}[theorem]{Corollary}
\fi

\ifx\definition\undefined
\newtheorem{definition}[theorem]{Definition}
\fi

\ifx\remark\undefined
\newtheorem{remark}[theorem]{Remark}
\fi

\def\x{{\mathbf{x}}}

\def\H{{\cal{H}}}

\newcommand{\sign}{\mathop{\rm sign}}

\newcommand{\RN}[1]{%
	\textup{\uppercase\expandafter{\romannumeral#1}}%
}
\def\clip{\textsf{clip}}
\def\FGSM{\textsf{FGSM}}
\def\KL{\textsf{KL}}
\ifx\remark\undefined
\newtheorem{remark}{Remark}
\fi

\begin{document}
%
\title{Distributionally Adversarial Attack}
\author{Tianhang Zheng\textsuperscript{1} \and Changyou Chen\textsuperscript{1} \and Kui Ren\textsuperscript{1, 2}\\
\textsuperscript{1} State University of New York at Buffalo\\ \textsuperscript{2} Zhejiang University\\
\{tzheng4, changyou, kuiren\}@buffalo.edu}

\maketitle
\begin{abstract}
Recent work on adversarial attack has shown that Projected Gradient Descent (PGD) Adversary is a universal first-order adversary, and the classifier adversarially trained by PGD is robust against a wide range of first-order attacks. It is worth noting that the original objective of an attack/defense model relies on a data distribution $p(\xb)$, typically in the form of risk maximization/minimization, {\it e.g.}, $\max\!/\!\min\mathbb{E}_{p(\xb)}\mathcal{L}(\xb)$ with $p(\xb)$ some unknown data distribution and $\mathcal{L}(\cdot)$ a loss function. However, since PGD generates attack samples independently for each data sample based on $\mathcal{L}(\cdot)$, the procedure does not necessarily lead to good generalization in terms of risk optimization. In this paper, we achieve the goal by proposing distributionally adversarial attack (DAA), a framework to solve an optimal {\em adversarial-data distribution}, a perturbed distribution that satisfies the $L_\infty$ constraint but deviates from the original data distribution to increase the generalization risk maximally. Algorithmically, DAA performs optimization on the space of potential data distributions, which introduces direct dependency between all data points when generating adversarial samples.
DAA is evaluated by attacking state-of-the-art defense models, including the adversarially-trained models provided by {\em MIT MadryLab}. Notably, DAA ranks {\em the first place} on MadryLab's white-box leaderboards, reducing the accuracy of their secret MNIST model to $88.79\%$ (with $l_\infty$ perturbations of $\epsilon = 0.3$) and the accuracy of their secret CIFAR model to $44.71\%$ (with $l_\infty$ perturbations of $\epsilon = 8.0$). Code for the experiments is released on \url{https://github.com/tianzheng4/Distributionally-Adversarial-Attack}. 
\end{abstract}
\section{Introduction}
Recent years have witnessed widespread use of deep neural networks (DNNs), achieving remarkable performance on different machine-learning tasks, such as object detection and recognition \citep{krizhevsky2012imagenet}, strategy optimization \citep{silver2016mastering}, and natural language processing \citep{cho2014learning}. At the same time, DNNs also have been proved to be vulnerable to adversarial samples -- data that are indistinguishable from natural samples by human but endow additional maliciously-embedded perturbations. Those maliciously perturbed samples can cause DNNs to make predictions different from the ground truth with high confidence. Various first-order algorithms have been proposed to generate adversarial samples, such as Fast Gradient Sign Method (FGSM) \citep{szegedy2013intriguing}, Projected Gradient Descent (PGD) \citep{kurakin2016adversarial}, and Carlini \& Wagner Attacks (CW) \citep{carlini2017towards}.

Among all those first-order attacks, \citet{madry2017towards} suggest that PGD is a universal attack algorithm, and the classifier adversarially trained by PGD is robust against a wide range of first-order attacks. \citet{nicholas2017ground} strengthen the hypothesis by demonstrating that PGD-adversarial training provably succeeds at increasing the distortion required to construct adversarial examples by a factor of 4.2. Moreover, among all the white-box defenses that appeared in ICLR-2018 and CVPR-2018, PGD-adversarial training is the only empirical defense that has not been further attacked \citep{athalye2018obfuscated,athalye2018robustness}.

Despite the success of PGD, one notable limitation is that the adversarial samples are not globally optimal, in the sense that adversarial samples are generated independently for each data sample. From a machine-learning perspective, this lacks a statistical interpretation in terms of risk maximization, {\it i.e.}, PGD is not a training procedure, thus the underlying optimization problem is not mathematically clear. In this paper, we provide a distribution-optimization view of PGD, and propose distributionally adversarial attack (DAA), a new concept of adversarial attack that is performed on the space of probability measures ({\it e.g.}, unknown data distributions). In DAA, the problem is formulated as optimizing an adversarial data distribution (from which adversarial samples are drawn from) such that the {\em generalization risk} increases maximumly. This generalizes PGD by lifting the optimization onto the space of probability measures, and can be interpreted as Wasserstein gradient flows (WGFs), a framework for distribution optimization which always decreases an ``energy functional'' over time. The energy functional reflexes the data manifold in adversarial attack, and is designed in correspondence with the original objective of a DNN. When using testing data to approximate the unknown data manifold, DAA leads to a variant of the standard PGD where all adversarial samples are explicitly dependent.



DAA is extensively evaluated on four datasets, including MNIST, Fashion MNIST (FMNIST), CIFAR10, and Imagenet, by attacking their state-of-the-art defense models. We show that a single run of DAA with $0.3/1.0$ $l_\infty$ perturbations can reduce the accuracy of MadryLab's MNIST model \citep{madry2017towards} to approximately $90.5\%$, outperforming a single run of PGD that reduces the accuracy to approximately $92.5\%$. Furthermore, a single run of DAA with $0.2/1.0$ $l_\infty$ perturbations reduces the accuracy of our adversarially trained FMNIST model to $67.6\%$, outperforming a single run of PGD that reduces the accuracy to $71.5\%$. Similarly, a single run of DAA reduces the accuracy of MadryLab's CIFAR10 model to $44.98\%$ ($8/255$ $l_\infty$ perturbations) and the accuracy of the ensemble adversarial trained Imagenet model \citep{kurakin2016adversarial} to $16.43\%$ (only $2/255$ $l_\infty$ perturbations). For DAA with $50$ random restarts, it reduces the accuracy of MadryLab's public and secret MNIST model to $88.7\%$ and $88.79\%$, respectively; whereas DAA with $10$ random restarts reduces the accuracy of MadryLab's secret CIFAR10 model to $44.71\%$. \textit{Both settings outperform other attack algorithms listed in MIT MadryLab's white-box leaderboards.}\footnote{\url{https://github.com/MadryLab/mnist_challenge}; \url{https://github.com/MadryLab/cifar10_challenge}}

\section{Preliminaries}
We introduce necessary background in this section, including Wasserstein gradient flows and adversarial attack/defense methods.
\subsection{Wasserstein Gradient Flows}
\paragraph{Wasserstein Metric Space}
Wasserstein metric is a distance metric defined between probability measures (distributions) on the Wasserstein metric space. Formally, let $P_2(\Omega)$ denote the collection of all probability measures on $\Omega \subset R^r$ with finite 2nd moment. The 2nd-order Wasserstein distance between two probability measures in $P_2(\Omega)$ is defined as:
\begin{align}
W^2_2(\mu, \nu) \triangleq \inf_\gamma\{\int_{\Omega \times \Omega} ||\xb - \xb^\prime||^2_2d\gamma(\xb, \xb^\prime):&
\\ \gamma \in \Omega(\mu, \nu)\}, \nonumber&
\end{align}
where $\Gamma (\mu ,\nu )$ denotes the collection of all joint probability measures on $\Omega \times \Omega$ with two marginals equal to $\mu$ and $\nu$.
One way to understand the motivation of the above definition is to consider the optimal transport problem, where one wants to transform elements in the domain of $\mu$ to $\nu$ with minimum cost. The cost to transport $\xb$ in $\mu$ to $\xb^\prime$ in $\nu$ is quantified by $||\xb - \xb^\prime||^2_2$. If $\mu$ is absolutely continuous w.r.t. the Lebesgue measure, there exists a unique optimal transport plan, i.e., a mapping $T : R^r \rightarrow R^r$, to transform elements in $\mu$ to elements in $\nu$. The Wasserstein distance can be equivalently reformulated as: $W^2_2(\mu, \nu) \triangleq \inf_T\{\int_{\Omega} ||\xb - T(\xb)||^2_2d\mu(\xb)$\}.


\paragraph{Wasserstein Gradient Flows}
The 2nd-order Wasserstein metric endows $P_2(\Omega)$ with a Riemannian geometry. Let $\{\mu_t\}_{t \in [0, 1]}$ be an absolutely continuous curve on this geometry, with change between $\mu_t$ and $\mu_{t+h}$ measured by $W^2(\mu_t, \mu_{t+h})$. The change can be reflected by a vector field: $v_t(\xb) \triangleq \lim_{h \rightarrow 0}\frac{T(\xb) - \xb}{h} $. This vector field is regarded as the velocity field of the elements $\xb$. Based on the above description, a gradient flow can be defined on $P_2(\Omega)$ in Lemma~\ref{lem:lem1}.
\begin{lemma}\label{lem:lem1}
	Let $\{\mu_t\}_{t \in [0, 1]}$ be an absolutely continuous curve in $P_2(\Omega)$. Then for a.e. $t \in [0, 1]$, the above vector field $\mathbf{v}_t$ defines a gradient flow on $P_2(\Omega)$ as: $\partial_t\mu_t + \nabla\cdot(\mathbf{v}_t \mu_t) = 0$.
\end{lemma}
Actually, the velocity field $\vb$ can be derived for optimization on an {\em energy functional} $E: P_2(\Omega) \rightarrow R$. In this case, it can be shown that $\mathbf{v}_t$ has the form $\mathbf{v}_t = -\nabla\frac{\delta E}{\delta \mu_t}$ \citep{ambrosio2008gradient}, where $\frac{\delta E}{\delta \mu_t}$ is called the first variation of $E$ at $\mu_t$. Thus the gradient flow on $P_2(\Omega)$ can be rewritten as:
\begin{equation}\label{eq:gradflow}
\partial_t\mu_t = - \nabla\cdot(\mathbf{v}_t \mu_t) = \nabla\cdot(\mu_t \nabla\frac{\delta E}{\delta \mu_t})
\end{equation}
\subsection{Adversarial Sample}
\label{sec:adv}
\paragraph{Definition and Notation} In this paper, we only study adversarial samples on neural networks used for classification, with final layers as softmax activation functions. We represent such a network as a vector function $\{F_i(\xb)\}_i$. Given an input $\xb$, the network predicts its label as $\tilde{y} = \arg_i\max F_i(x)$. A sample $\xb^\prime$ is called an adversarial sample if $\arg_i\max F(\xb^\prime) \neq y$, where $y$ is the true label and $\xb^\prime$ is close to the original $\xb$ under a certain distance metric.

\paragraph{Fast Gradient Sign Method (FGSM)}
Fast Gradient Sign Method (FGSM) is a single-step adversarial attack proposed by \citet{szegedy2013intriguing}. FGSM performs a single step update on the original sample $\xb$ along the direction of the gradient of a loss function $\mathcal{L}(\xb, y; \thetab)$. The loss function is usually defined as the cross-entropy between the output of a network and the true label $y$.
Formally, FGSM adversarial samples are generated as
\begin{equation}\label{eq:fgsm}
\xb^\prime = \clip_{[0, 1]}\{ \xb + \epsilon \cdot \sign(\nabla_{x} \mathcal{L}(\xb, y; \thetab)\}~,
\end{equation}
where $\epsilon$ controls the maximum $l_\infty$ perturbation of the adversarial samples, and the $\clip_{[a,b]}(\xb)$ function forces $\xb$ to reside in the range of $[a, b]$.

\paragraph{Projected Gradient Descent (PGD)}
Projected Gradient Descent (PGD) is an iterative variant of FGSM. In each iteration, PGD follows the update rule:
\begin{equation}\label{eq:pgd}
\xb^\prime_{l+1} = \RN{2}_{\clip}\{\FGSM(\xb^\prime_l)\}~,
\end{equation}
where $\FGSM(\xb^\prime_l)$ represents an FSGM update of $\xb^\prime_l$ as in \eqref{eq:fgsm}, and the outer clip function $\RN{2}_{\clip}$ keeps $\xb^\prime_{l+1}$ within a predefined perturbation range.
PGD can also be interpreted as an iterative algorithm to solve the following problem:
\begin{equation}\label{eq:advProb}
\max_{\xb^\prime: ||\xb^\prime-\xb||_\infty < \alpha} \mathcal{L}(\xb^\prime, y; \thetab).
\end{equation}
\citet{madry2017towards} observe that the local maxima of the cross-entropy loss found by PGD with $10^5$ random starts are distinctive, but all have similar loss values, for both normally- and adversarially-trained networks. Inspired by this concentration phenomena, they propose that PGD is a universal adversary among all the first-order adversaries, {\it i.e.}, attacks only rely on first-order information.

\paragraph{Momentum-based Iterative Fast Gradient Sign Method (MI-FGSM)}
MI-FGSM is derived from the Iterative FGSM \citep{kurakin2016physical}, which integrates the momentum term into an iterative process to generate adversarial samples \citep{dong2018boosting}. Given $g_0 = 0$ and $g_{l+1} = \mu \cdot g_l + \frac{\nabla_{x} \mathcal{L}(\xb'_l, y; \thetab)}{||\nabla_{x} \mathcal{L}(\xb'_l, y; \thetab)||_1}$, the iterative version of MI-FGSM can be expressed as:
\begin{equation}\label{eq:mifgsm}
\xb'_{l+1} = \xb'_l + \epsilon \cdot \sign(g_{l+1}).
\end{equation}
Based on MI-FGSM, we further derive its PGD variant, called Momentum PGD, with iterative updates
\begin{equation}\label{eq:mifgsm1}
\xb'_{l+1} = \xb'_l + \cdot \RN{2}_{\clip}\{\epsilon \cdot \sign(g_{l+1})\}.
\end{equation}
\begin{remark}
	Momentum PGD is a stronger attack than MI-FGSM, since it can proceed for more steps with an appropriate step size $\epsilon$ ($\epsilon$ can not be too small, otherwise the adversarial samples are very likely to get trapped in bad local maxima). The clip function $\RN{2}_{\clip}$ ensures the adversarial samples to have the predefined perturbation size after the extra iterations.
\end{remark}

\subsection{Adversarial Training}
\paragraph{Definition and Notation} Adversarial training is a defense method against adversarial samples first proposed by \citet{goodfellow2014}. The approach attempts to improve the robustness of a network by training it together with adversarial samples. Formally, adversarial training solves the following min-max problem:
\begin{equation}\label{eq:advTrain}
\min_\theta \max_{x': D(\xb, \xb') < \alpha} \mathcal{L}(\xb', y; \thetab)~,
\end{equation}
where $D(\xb, \xb')$ represents certain distance metric between $x$ and $x'$. The inner maximization problem is equivalent to constructing the strongest adversarial samples. If $l_\infty$ distance is employed as the distance metric $D(\xb, \xb')$, the inner maximization problem is equivalent to the adversarial problem solved by PGD, i.e., \eqref{eq:advProb}. The outer minimization is the standard training procedure to minimize the loss of a DNN. Recent work shows that this straightforward method is one of the most effective defenses against adversarial samples \citep{madry2017towards,Alexey2017Ensemble,cai2018curriculum}.

\paragraph{PGD Adversarial Training}
The fact that PGD adversary is a first-order universal adversary implies that robustness against PGD should yield robustness against all first-order adversaries \citep{madry2017towards}. Hence, \citet{madry2017towards} propose to adversarially train a robust classifier using PGD attack. Specifically, in each training iteration, PGD is applied to generate a minibatch of adversarial samples to update the current network. In the training process, a steady decrease of the training loss is usually observed, indicating the effectiveness of this training paradigm.
Experiment results show that PGD adversarial-trained models are robust against PGD attack as well as another strong attacks such as the $CW_\infty$ attack \citep{carlini2017towards}. Empirically we also found that their MNIST and CIFAR-10 models are indeed robust to a wide range of existing first-order attacks, including DeepFool \citep{moosavi2016deepfool}, and Jacobian-based Saliency Map Attack (JSMA) \citep{papernot2016limitations}, as long as the adversarial perturbations are $l_\infty$- bounded. 

\paragraph{Ensemble Adversarial Training}
To scale up adversarial training to ImageNet-scale datasets, \citet{kurakin2016adversarial} adversarially train a model using a fast single-step attack method. However, their adversarially-trained model is vulnerable to multi-step white-box attacks \citep{kurakin2016adversarial}. \citet{Alexey2017Ensemble} further demonstrate that the model of \citet{kurakin2016adversarial} is vulnerable to black-box adversaries \citep{Alexey2017Ensemble}. To tackle this problem, \citet{Alexey2017Ensemble} propose a training methodology that incorporates adversarial samples transferred from other pre-trained models, called Ensemble Adversarial Training (EAT) \citep{Alexey2017Ensemble}. Intuitively, this approach increases the diversity of adversarial samples used for adversarial training. In their experiments, the models trained by EAT exhibit robustness against adversarial samples transferred from other holdout models, using various single-step and multi-step attacks.

\section{Distributionally Adversarial Attack}
We first interpret distributionally adversarial attack as WGFs, and then propose a specific energy functionals to construct a WGF for better adversarial-sample generation, and finally propose particle-approximation methods to solve the DAA problem, leading to a variant of the standard PGD.
\subsection{Adversarial Attack as WGFs}
For a given DNN, the landscape of a loss function $\mathcal{L}(\xb, y; \thetab)$ constitutes a geometry structure indexed by input images $\xb$. From a probability perspective, under regularized conditions, it is natural to define a probability distribution for each input $\xb$ based on the loss-function landscape, {\it i.e.},
\begin{align}
p(\xb, y; \thetab) \propto \exp\{-\mathcal{L}(\xb, y; \thetab)\}~.
\end{align}
Since $y$ is deterministic given $\xb$, we would omit $y$, and write $p(\xb, y; \thetab)$ as $p(\xb; \thetab)$ in the following for simplicity. Based on this, we explain our intuitions on generalizing adversarial attack on the space of data distributions in the following. First note that the objective of an adversarial attack \eqref{eq:advProb} is equivalently rewritten as:
\begin{align}\label{eq:aa}
\xb^\prime = \arg\max_{x': D(x, x') < \alpha}\{\mathcal{L}(\xb^\prime, y; \thetab) - \mathcal{L}(\xb, y; \thetab)\}~,
\end{align}
which describes the increase of a loss with an adversarial sample. On the space of probability measure, the loss is instead described by an energy functional $E(\mu)$, assuming the minimum being reached at $p(\xb; \thetab)$. Consequently, instead of finding an optimal adversarial sample $\xb^\prime$ for each $\xb$, DAA tries to find an optimal {\em adversarial-data distribution}, $\mu^{*}$, such that $\mu^{*}$ is close to $p(\xb; \thetab)$ but increases $E(\cdot)$ maximumly, {\it i.e.},
\begin{align}\label{eq:daa}
\mu^{*} &= \arg\max_{\mu: W_2(\mu, p(\xb; \thetab)) < \alpha}\{E(\mu) - E(p)\} \nonumber\\
&= \arg\max_{\mu: W_2(\mu, p(\xb; \thetab)) < \alpha}E(\mu)~.
\end{align}

\begin{theorem}
	The solution $\mu^{*}$ of \eqref{eq:daa} is equivalently described by the following PDE:
	\begin{align}\label{eq:daawgf}
	\partial_t \mu_t = -\nabla_{\xb}\cdot \left(\mu_t\nabla_{\xb}\left(\frac{\delta E}{\delta\mu_t}(\mu_t)\right)\right)~,
	\end{align}
	and $\mu^{*} = \mu_t$ where $t = \{\inf\{t^\prime\}: \mu_0(\xb) = p(\xb; \thetab), W_2(\mu_{t^\prime}(\xb), p(\xb; \thetab)) < \alpha\}$.
\end{theorem}

\subsection{Energy Functional}
It is crucial to define an energy functional as it directly affects adversarial-sample behaviors. Recent studies \citep{song2017pixeldefend,ma2018characterizing} show that adversarial samples/subspaces mainly lie in the low probability regions of the original data distribution, thus we expect adversarial distribution to derivate from the original distribution by optimization over the energy functional. 
Besides, the energy functional should also be simple enough to possess a unique solution, {\it i.e.}, it should be convex w.r.t.\! $\mu$ on the space of probability measures.Therefore, we define a new energy functional as:
\begin{align}\label{eq:energy}
E(\mu) = \int_{\mu} \mathcal{L}(\xb, y; \thetab) d\mu + c \cdot \KL(\mu||p)~,
\end{align}
where $\mathcal{L}(\xb, y; \thetab)$ is the loss of the system; $KL(\mu||p)$ is the KL-divergence between the adversarial distribution $\mu$ and the optimal data distribution $p$; and $c$ is a hyperparameter balancing those two terms. Intuitively, maximizing \eqref{eq:energy} will increase the individual losses in addition to the deviation between the adversarial and original distributions. Note the energy functional \eqref{eq:energy} is still convex w.r.t.\! $\mu$, maintaining the optimality condition and making the problem easier to solve.

\subsection{Adversarial-Distribution Optimization and Adversarial-Sample Generation}\label{sec:generation}
Note a closed-formed solution of \eqref{eq:daawgf} is infeasible given the energy functional defined by \eqref{eq:energy}. Following standard methods such as those in \citep{chen2018unified}, we adopt particle approximation to solve \eqref{eq:daawgf}. The idea is to approximate $\mu$ with a set of $M$ particles $\{\xb^{(i)}\}_{i=1}^M$ as
$\mu \approx \frac{1}{M}\sum_{i=1}^M\delta_{\xb^{(i)}}$, 
where $\delta_{\xb}$ is a delta function with a spike at $\xb$. Consequently, solving for the optimal $\mu$ corresponds to optimizing the particles as {\em adversarial samples} from the adversarial distribution.
In the following, based on \citep{chen2018unified}, we investigate two methods for particle approximation: the Lagrangian blob method and the discrete-gradient-flow method.
\paragraph{Lagrangian Blob Method}
The idea is to use particle approximations directly in the original problem \eqref{eq:daawgf}. Specifically, define $\mathbf{v}_t \triangleq \nabla_{\xb}\left(\frac{\delta E}{\delta\mu_t}(\mu_t)\right)$. According to \citep{carrillo2017blob}, $\vb_t$ is interpreted as the velocity function of a particle in the gradient flow. Consequently, Lagrangian blob methods evolve particles on a grid with a time-spacing $h$ following the velocity $\vb_t$ \citep{carrillo2017blob}. Thus solving the WGF \eqref{eq:daawgf} is equivalent to evolving the particles along their velocities as $$d\xb^{(i)}/dt = \mathbf{v}_t(\xb^{(i)})~.$$
To calculate $\vb_t$, we substitute the form of $E(\mu)$ in \eqref{eq:energy} into $\vb_t$. First note that under the $\H$-Wasserstein distance metric defined by \citep{liu2017Stein}, we have
\begin{align}\label{eq:svgd}
&\nabla_{\xb}\left(\frac{\delta KL(\mu_t|p)}{\delta\mu_t}\right) = \mathbb{E}_{\tilde\xb \sim \mu_t}[\nabla_{\tilde\xb}[p(\tilde\xb)K(\xb, \tilde\xb)]/p(\tilde\xb)] \nonumber\\
=& \mathbb{E}_{\tilde\xb \sim \mu_t}[K(\xb, \tilde\xb)\nabla_{\tilde\xb}\log p(\tilde\xb) + \nabla_{\tilde\xb} K(\xb, \tilde\xb)],
\end{align}
where $K(\cdot, \cdot)$ is a kernel function such as the RBF kernel.
For the first term on the right of \eqref{eq:energy}, we have
\begin{multline}\label{eq:gradloss}
\nabla_{\xb}\left(\frac{\delta \int_{\mu} \mathcal{L}(\xb, y; \thetab) d\mu }{\delta\mu_t}\right) = \nabla_{\xb} \mathcal{L}(\xb, y; \thetab).
\end{multline}
Combining Eq. \ref{eq:svgd} and \ref{eq:gradloss}, one ends up solving the following ordinary differential
equation:
\begin{align}\label{eq:daa_svgd1}
\frac{d\xb}{dt} &= \nabla_{\xb} \mathcal{L}(\xb, y; \thetab) + \\
& c \cdot \mathbb{E}_{\tilde\xb \sim \mu_t}[K(\xb, \tilde\xb)\nabla_{\tilde\xb}\log p(\tilde\xb) + \nabla_{\tilde\xb} K(\xb, \tilde\xb)].\nonumber
\end{align}
Considering a discrete approximation of $\mu_t$ with particles, \eqref{eq:daa_svgd1} can be solved numerically as
\begin{align}\label{eq:daa_svgd}
\xb_{\ell+1}^{(i)} = \xb_{\ell}^{(i)} + \epsilon_l\cdot\{\nabla_{\xb_{\ell}^{(i)}} \mathcal{L}(\xb_{\ell}^{(i)}, y^{(i)}; \thetab) + ~~~~~&\\
 \frac{c}{M} \sum_{j=1}^M[K(\xb_{\ell}^{(i)}, \xb_{\ell}^{(j)})\nabla_{\xb_{\ell}^{(j)}}\log p(\xb_{\ell}^{(j)}) + & \nonumber\\
\nabla_{\xb_{\ell}^{(j)}} K(\xb_{\ell}^{(i)}, \xb_{\ell}^{(j)})]\},& \nonumber
\end{align}
where, in contrast to the continuous case, we use $\ell$ to index the number of steps for the discretized particles.

\paragraph{Discrete Gradient Flows}
Discrete-gradient-flow (DGFs) approximation for \eqref{eq:daawgf} consists of a sequence of sub-optimization problems whose composition approximates $\mu_t$, {\it i.e.}, $\mu_t \approx \tilde{\mu}_{L}\circ\cdots\circ \tilde{\mu}_1$, where $L = t / h$, and $\tilde{\mu}_{\ell}$ is the solution of the following functional optimization problem \footnote{Note the difference between \eqref{eq:wgf} and the original DGF formula is to replace the original $\min$ to $\max$ because the flow direction is reversed in adversarial-distribution optimization.}:
\begin{align}\label{eq:wgf}
\tilde{\mu}_{\ell} = \arg \max_{\mu \in P_2(R^r)} \{E(\mu) - \frac{1}{2h}W^2_2(\tilde{\mu}_{\ell-1}, \mu)\}.
\end{align}
Again, \eqref{eq:wgf} is solved by particle approximation, with gradient ascent on the particles. To this end, we need gradients for the two terms on the RHS of \eqref{eq:wgf}. Inspired by \citep{chen2018unified}, we decompose the two terms and re-organize terms:
\begin{align*}
&E_1 \triangleq \sum_{i=1}^M \left(\mathcal{L}(\xb_{\ell}^{(i)}, y^{(i)}; \thetab) - c \log p(\xb_{\ell}^{(i)})\right) \\
&E_2 \triangleq \mathbb{E}_{\mu}[\log \mu] + \frac{1}{2h}W_2^2(\mu, \tilde{\mu}_{\ell-1})~.
\end{align*}
The gradient of the first term can be easily calculated as
\begin{align}\label{eq:deriveE1}
\frac{\partial E_1}{\partial \xb_{\ell}^{(i)}} &= \nabla_{\xb_{\ell}^{(i)}}\mathcal{L}(\xb_{\ell}^{(i)}, y^{(i)}; \thetab) + c \nabla_{\xb_{\ell}^{(i)}}\log p(\xb_{\ell}^{(i)}) \nonumber\\
&= (1 + c)\nabla_{\xb_{\ell}^{(i)}}\mathcal{L}(\xb_{\ell}^{(i)}, y^{(i)}; \thetab)~.
\end{align}
For the $E_2$ term, we apply similar idea as \citep{chen2018unified} by introducing Lagrangian multipliers, resulting in

\begin{equation}\label{eq:daa_sgld}
\frac{\partial E_2}{\partial \xb^{(i)}_{\ell}} \approx c \cdot [\sum_{j}2u_iv_j(\frac{d_{ij}}{\lambda} - 1)e^{-\frac{d_{ij}}{\lambda}}(\xb^{(i)} - \xb^{(j)}_{k-1})],
\end{equation}
where $\lambda$, $u_i$ and $v_j$ are Lagrangian multipliers, and $d_{ij} = ||\xb^{(i)} - \xb^{(j)}_{k-1}||_2^2$. For the sake of simplicity, we do not update $u_i$ and $v_j$, but instead use a fixed scaling factor $\gamma$ to approximate the product $u_iv_j$.

\paragraph{Adversarial-Sample Generation}
Once an {\em adversarial distribution} is learned, adversarial samples, {\it e.g.} $\xb_{\ell}^{(i)}$, can be obtained by drawing samples from it. However, adversarial samples typically follow certain constraints, {\it e.g.}, $l_{\infty}$ bounded. We propose two adversarial-sample generation methods based on the particle-optimization formula above, named DAA-BLOB and DAA-DGF. DAA-BLOB substitutes the gradient used in PGD with the gradient derived in Eq. \ref{eq:daa_svgd}. Formally, in each iteration, $\xb^i$ is updated by
\begin{align}\label{eq:discretesvgd}
	\xb^{(i)}_{l+1} =  \Pi_{\clip}\{ \xb^{(i)}_l + \epsilon \cdot \sign(\nabla_{\xb^i_l} \mathcal{L}(\xb^{(i)}_l, y^{(i)}; \thetab) + &
	 \\ \frac{c}{M}[\sum_{j = 1}^{M}K(\xb^{(i)}_l, \xb^{(j)}_l)\nabla_{\xb^{(j)}_l}\mathcal{L}(\xb^{(j)}_l, y^{(j)}; \thetab) + & \nonumber\\  \nabla_{\xb^{(j)}_t}K(\xb^{(i)}_l, \xb^{(j)}_l)])\}.& \nonumber
\end{align}
By contrast, DAA-DGF substitutes the gradient used in PGD with a combination of Eq. \ref{eq:deriveE1} and Eq. \ref{eq:daa_sgld}. Specifically, in each iteration, $\xb^i$ is updated by
\begin{equation}
\begin{split}\label{eq:discretesgld}
\xb^{(i)}_{l+1} = & \Pi_{\clip}\{ \xb^{(i)}_l + \epsilon \cdot \sign(\nabla_{\xb^{(i)}_l} \mathcal{L}(\xb^{(i)}_l, y^{(i)}; \thetab) - \\
&\frac{2\gamma c}{1+c} \cdot [\sum_{j=1}^{M}(\frac{d_{ij}}{\lambda} - 1)e^{-\frac{d_{ij}}{\lambda}}(\xb^{(i)}_l - \xb^{(j)}_l)])\}.
\end{split}
\end{equation}

\paragraph{Optimization by Data Subsampling}
In theory, the adversarial distribution $\mu$ to be optimized corresponds to a data manifold. Thus a good discrete approximation to $\mu$ is to use all the testing samples.
In practice, however, it is computationally infeasible to update the particles following \eqref{eq:discretesvgd} or \eqref{eq:discretesgld}, as the complexity is $\mathcal{O}(M^2)$ for each particle update. To mitigate this issue, we propose a subsampling method to update testing samples in an unbiased and computationally feasible way: First, testing samples are randomly permuted and divided into finite number of minibatches; then samples in each minibatch are updated sequentially for a certain number of steps. This procedure is iterated for multiple rounds. The full algorithm is shown in Algorithm \ref{alg:daa}.

\begin{algorithm}
	\caption{DAA algorithm (untargeted attack)}
	\label{alg:daa}
	\begin{algorithmic}
		\REQUIRE A classifier with loss function $\mathcal{L}(\xb^i, y^i; \thetab)$; testing dataset $\{\xb^i, y^i\}_{i=1}^{N}$;
		minibatch size $M$; step size $\epsilon$; predefined final perturbation size $\alpha$; total iterations $L$; rounds $R$; hyperparameter $c$ or $\frac{2\gamma c}{1+c}$;
		
		\STATE \textbf{Random Start}: $\xb_0^i = \xb^i + \boldsymbol\gamma^i$ ($\boldsymbol\gamma^i \sim \mathcal{U}(-\alpha, \alpha)$)
		\STATE \textbf{No Random Start}: $\xb_0^i = \xb^i$
		\FOR {$r$ = $0$ to $R-1$}
		\STATE Randomly permutate the testing samples
		\FOR {$k$ = $0$ to $L/R-1$}
		\STATE $l = rL/R + k$
		\FOR {$j$ = $0$ to $N/M-1$}
		\STATE Follow Eq. \ref{eq:discretesvgd} or \ref{eq:discretesgld} to update the minibatch $\{\xb_l^i, y_l^i\}$ ($i=jM+1 \sim (j+1)M$), where $\RN{2}_{\clip}(\cdot) = \clip_{[0, 1]}(\clip_{[\xb^i - \alpha, \xb^i + \alpha]} (\cdot))$  ($[0, 1]$ is the pixel value range, maybe $[-1, 1]$ or $[0, 255]$)
		\ENDFOR 
		\ENDFOR
		\ENDFOR
	\end{algorithmic}
\end{algorithm}

\paragraph{Connection with PGD}
There are two situations where the proposed DAA framework reduces to PGD. The first situation is when $c=0$, where the second terms of both \eqref{eq:discretesvgd} and \eqref{eq:discretesgld} become $0$, making the two gradients degraded to the gradient used in PGD. The second case is when $M=1$, meaning that DAA is equivalent to PGD when the data-manifold is approximated by only one sample. In this case, the second term of the gradient used in DAA-DGF becomes $0$, and the second term of the gradient used in DAA-BLOB reduces to $ c \cdot [K(\xb^i_t, \xb^i_t)\nabla_{\xb^i_t}\mathcal{L}(\xb^i_t, y^i; \thetab) + \nabla_{\xb^j_t}K(\xb^i_t, \xb^i_t)] = c \cdot \nabla_{\xb^i_t}\mathcal{L}(\xb^i_t, y^i; \thetab)$. 

\section{Experiment}
\begin{figure*}[!t]
	\centering
	\includegraphics[width=5in]{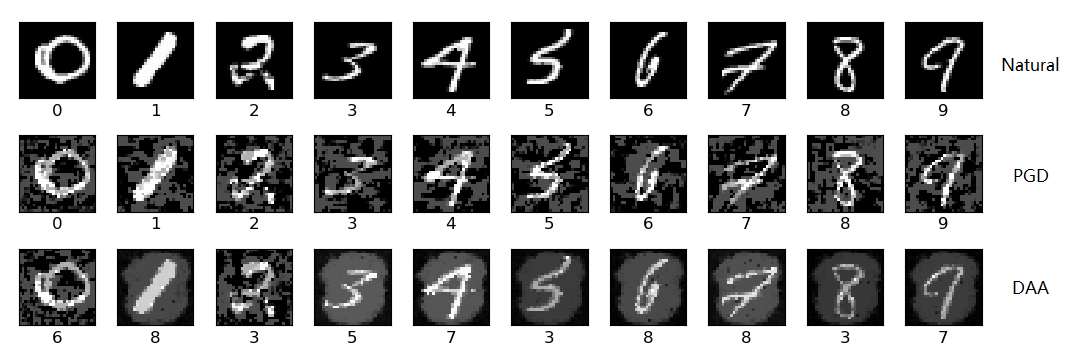}
	\caption{Comparison between PGD and DAA. DAA tends to generate more structured perturbations.}
	\label{fig:DAAexample}
\end{figure*}

\begin{table*}
	\caption{Empirical worst-case accuracy of MIT MadryLab's secret \textbf{MNIST} model under 200-step attacks with 50 random starts ($0.3/1.0$ $l_\infty$ perturbations). Loss 1: cross-entropy, Loss 2: $CW_\infty$ loss}
	\center
	\begin{tabular}{ |c|c|c|c|c|c|c|c|c|c| }
			\hline
			\multicolumn{2}{|c|}{Rand+FGSM}& \multicolumn{2}{|c|}{PGD}& \multicolumn{2}{|c|}{Momentum PGD}&\multicolumn{2}{c|}{DAA-BLOB}&\multicolumn{2}{c|}{DAA-DGF}\\
			\hline
			Loss 1 & Loss 2 & Loss 1 & Loss 2 & Loss 1 & Loss 2 & Loss 1 & Loss 2 & Loss 1 & Loss 2\\
			\hline
			$93.48\%$ & $93.47\%$ & $89.49\%$ & $89.57\%$ &  $89.29\%$ & $89.36\%$ & $\mathbf{88.79}\%$ & $\mathbf{88.85}\%$ & $\mathbf{88.92}\%$ & $89.25\%$ \\
			\hline
	\end{tabular}
	\label{tab:mnist}
\end{table*}

\subsection{Setup}
\paragraph{Datasets and Related Models}
The proposed DAA together with state-of-the-art methods, PGD and Momentum PGD, are evaluated and compared on four standard datasets, including MINST, Fashion MNIST (FMNIST), CIFAR10 and ImageNet. For MINST, the attack target is the state-of-the-art PGD-adversarially-trained MINST model provided by MIT MadryLab \citep{madry2017towards}. The defense architecture contains a convolutional neural network (CNN) with two convolutional layers and a fully connected layer. For FMNIST, we adversarially train a model by PGD as the target model. The network architecture consists of four convolutional layers and a fully connected layer with batch normalization.
For CIFAR10, MadryLab's PGD adversarially trained CIFAR10 model is adopted as the target model.
The network architecture is a residual CNN consisting of five residual units and a fully connected layer. For ImageNet, we adopt the target model in \citep{Alexey2016Adversarial}, which is an adversarially trained Inception ResNet-v2 model. 
\textit{In addition, we also evaluate DAA on a provable defense model \citep{wong2018provable}, with code also provided in the Github link (in abstract)}

\paragraph{Implementation Details} 
For all the methods related to kernel functions, an RBF kernel $K(\xb, \xb') = \exp(-\|\xb-\xb^\prime\|_2^2/h)$ is adopted. The bandwidth is set as $h = med^2/\log M$, same as the kernel used in \citep{liu2016Stein} and \citep{chen2018unified}. Here $med$ is the median of the pairwise distance between particles. The minibatch size (number of particles) is set to $100 \sim 200$ for computational feasibility. Our specific settings on hyper-parameters $c$ and $\frac{2\gamma c}{1+c}$ can be found in our Github link (in abstract). It is worth noting that the discrepancy regarding the parameter choices on those datasets is caused by different pixel ranges and network structures used by their classifiers. All experiments are conducted on a single Titan V GPU under a white-box setting, where an adversary has full access to a target model including model weights.

\subsection{Adversarial Perturbation Analysis}
To intuitively understand the advantage of DAA over PGD, we plot ten natural samples and their PGD and DAA adversarial samples in Figure~\ref{fig:DAAexample}. For the ten samples, DAA successfully attacks the defense model with a $0.3/1.0$ $l_\infty$ perturbation, whereas PGD with 50 random starts cannot. As shown in Figure~\ref{fig:DAAexample}, the perturbations generated by PGD tend to scatter throughout the images, whereas those of DAA are more structuredly focused around the target digits.

\subsection{Empirical Results}
\begin{table*}
	\caption{Empirical worst-case accuracy of adversarially trained \textbf{FMNIST} model under 100-step attacks with 10 random starts ($0.2/1.0$ $l_\infty$ perturbations). Loss 1: cross-entropy, Loss 2: $CW_\infty$ loss}
	\center
	\begin{tabular}{ |c|c|c|c|c|c|c|c|c|c| }
			\hline
			\multicolumn{2}{|c|}{Rand+FGSM}& \multicolumn{2}{|c|}{PGD}& \multicolumn{2}{|c|}{Momentum PGD}&\multicolumn{2}{c|}{DAA-BLOB}&\multicolumn{2}{c|}{DAA-DGF}\\
			\hline
			Loss 1 & Loss 2 & Loss 1 & Loss 2 & Loss 1 & Loss 2 & Loss 1 & Loss 2 & Loss 1 & Loss 2\\
			\hline
			$77.45\%$ & $77.21\%$ & $68.54\%$ & $68.94\%$ &  $69.72\%$ & $69.51\%$ & $\mathbf{65.70}\%$ & $66.64\%$ & $66.04\%$ & $66.60\%$ \\
			\hline
	\end{tabular}
	\label{tab:fmnist}
\end{table*}
\begin{table*}
	\caption{Empirical worst-case accuracy of MIT MadryLab's adversarially trained \textbf{CIFAR10} model under a single run of 100-step attacks without random start (8, 16/255 $l_\infty$ perturbations). Loss 1: cross-entropy, Loss 2: $CW_\infty$ loss}
	\center
	\begin{tabular}{ |c|c|c|c|c|c|c|c|c|c|c| }
			\hline
			$l_\infty$&\multicolumn{2}{|c|}{Rand+FGSM}&\multicolumn{2}{|c|}{PGD}&\multicolumn{2}{|c|}{Momentum PGD}&\multicolumn{2}{c|}{DAA-BLOB}&\multicolumn{2}{c|}{DAA-DGF}\\
			\cline{2-11}
			&Loss 1 & Loss 2 &Loss 1 & Loss 2 & Loss 1 & Loss 2 & Loss 1 & Loss 2 & Loss 1 & Loss 2\\
			\hline
			$8/255$ &$55.63\%$ & $55.05\%$ &$45.09\%$ & $46.27\%$ & $45.86\%$ & $46.76\%$ & $\mathbf{44.98}\%$ & $46.30\%$ & $45.07\%$ & $46.31\%$ \\
			\hline
			$16/255$ & $38.80\%$ & $37.78\%$ & $14.59\%$ & $16.06\%$ & $17.73\%$ & $18.70\%$ & $\mathbf{14.43}\%$ & $16.05\%$ & $14.52\%$ & $16.06\%$ \\
			\hline
	\end{tabular}
	\label{tab:cifar}
\end{table*}
\begin{figure}
	\includegraphics[width=0.48\columnwidth]{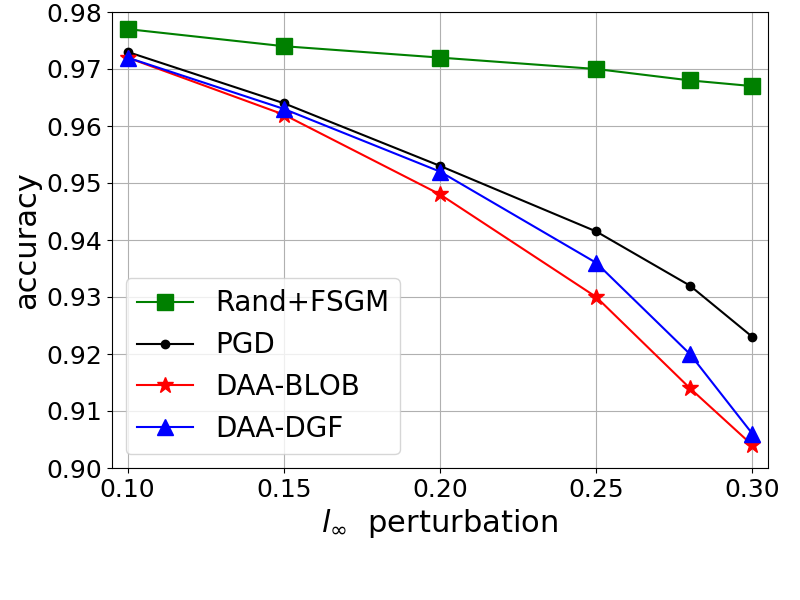}
	\includegraphics[width=0.48\columnwidth]{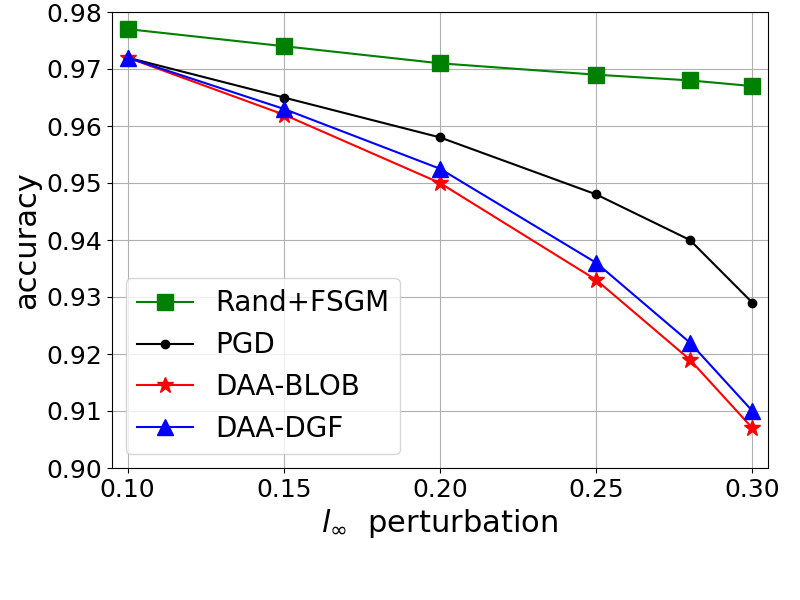}
	\caption{Averaged classification accuracy of MIT MadryLab's adversarially trained MNIST model under a single run of different attacks: cross-entropy (left), $CW_\infty$ loss (right)}
	\label{fig:mnist}
\end{figure}
\paragraph{MNIST}
We plot the averaged classification accuracy of MadryLab's adversarially trained MNIST model under a single run of different attack algorithms in Figure~\ref{fig:mnist}. It is observed that the proposed DAA consistently outperforms other methods under different levels of $l_{\infty}$ perturbations and two different losses. To test its statistic significance, we also conduct paired t-tests between the accuracies reduced by PGD and DAA with random starts. For DAA-BLOB and DAA-DGF, the $p$-values are almost zeros, {\it i.e.}, $0.0$ for DAA-BLOB and 5e-43 for DAA-DGF in the given decimal degree accuracy,
suggesting that both methods outperform PGD in a statistical sense. In addition, we show the worst classification accuracy of Madry's adversarially trained model under PGD, Momentum PGD and DAA with $50$ random restarts in Table~\ref{tab:mnist}. It is seen that DAA-BLOB is able to reduce the classification accuracy to approximately $88.79\%$ (with $c=1.1$ and minibatch size of 200), outperforming the attacks listed in MadryLab's white-box leaderboard.

\begin{figure}
	\includegraphics[width=0.48\columnwidth]{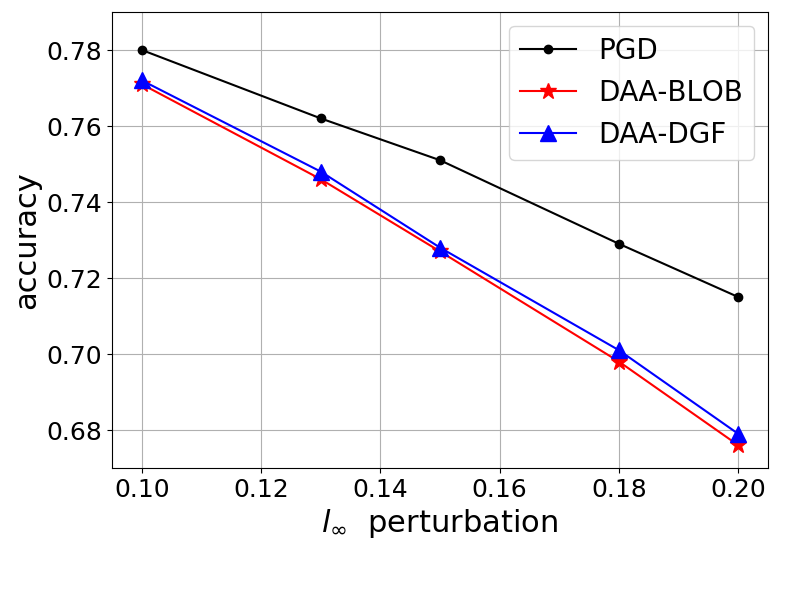}
	\includegraphics[width=0.48\columnwidth]{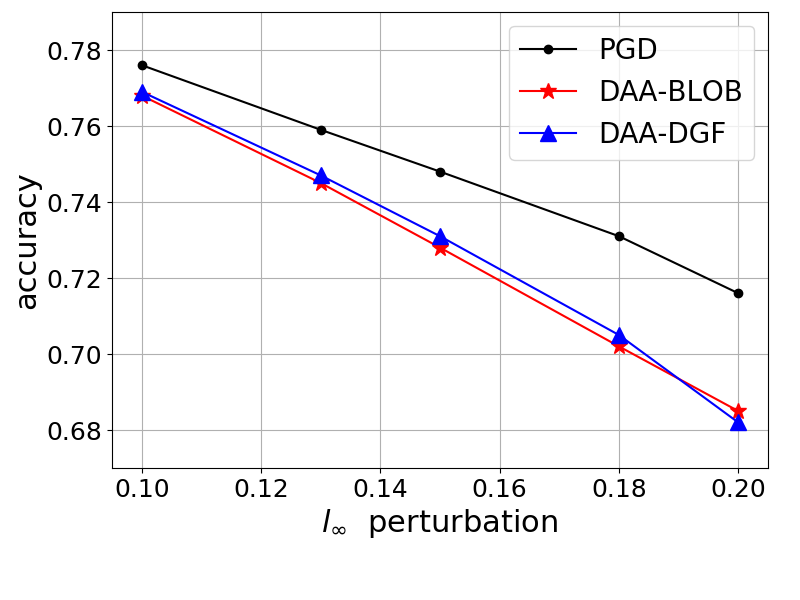}
	\caption{Averaged classification accuracy of adversarially trained FMNIST model under a single run of different attacks: cross-entropy (left), $CW_\infty$ loss (right)}
	\label{fig:fmnist}
\end{figure}

\paragraph{FMNIST}
We next plot the classification accuracy of our adversarially trained FMNIST model under a single run of different attack algorithms in Figure~\ref{fig:fmnist}. Similarly, the proposed DAA-based methods outperform the state-of-the-art PGD method. The two variants, DAA-BLOB and DAA-DGF, are comparable. Again, the $p$-values in the $t$-tests are also close to zero, {\it i.e.}, $0.0$ for DAA-BLOB and 3e-27 for DAA-DGF, indicating significant performance differences between PGD and the proposed methods. 
The worst accuracy under white-box PGD, Momentum PGD and DAA with $10$ random starts are listed in Table \ref{tab:fmnist}, suggesting the advance of the proposed DAA framework over both PGD and Momentum PGD.

\paragraph{CIFAR10}
The classification accuracies of Madry's adversarially trained model under a single run of PGD, Momentum PGD and DAA without random start are shown in Table. \ref{tab:cifar}. As can be seen, the adversarially-trained model only achieves weak robustness, {\it i.e.}, a 100-step PGD with $16/255$ $l_\infty$ perturbations reduces the accuracy to $14.59\%$, while DAA only reduces it to $14.43\%$. We conjecture this is because the data are too complex and sparse, making the particle approximation with testing samples in DAA badly represent the true data manifold. As a result, testing results with the learned adversarial samples are similar to those of PGD. This argument is also validated by \citet{recht2018cifar}, which shows that existing high-accuracy CIFAR10 classifiers does not generalize well to a truly unseen CIFAR10 testing set.

\paragraph{ImageNet}

We also evaluated the ensemble adversarial trained Inception ResNet-v2 under 50-step targeted Rand+FGSM and DAA attacks, using the least likely class as the target. Our experiments show that Rand+FGSM with {\em $\mathbf{2/255}$ $l_\infty$ perturbations} (approximately 0.0157/2.0 $l_\infty$ perturbations after normalization) can reduce the accuracy of the Inception model to approximately $70\%$ \citep{kurakin2016adversarial}, while 50-step DAA can dramatically reduce it to $16.43\%$. This indicates the ensemble adversarial trained ImageNet model is still vulnerable to well-designed iterative attacks, {\it e.g.,} our DAA.

\section{Discussion}
To our knowledge, there was not a first-order $l_\infty$ attack algorithm that can really outperform PGD under the white-box setting. Even for MI-FGSM, which won the NIP2017 competition under a black-box setting, we found that its PGD variant, which is stronger than MI-FGSM, cannot outperform standard PGD under the white-box setting, let alone MI-FGSM. In this paper, we generalize PGD on the space of data distributions. Our theoretical derivations and experiments validate the effectiveness of our framework. To the best of our knowledge, the proposed DAA framework is the first-and-only first-order $l_\infty$ attack algorithm that can outperform PGD, especially on robust adversarially trained models. To further attack those adversarially-trained models with small $l_\infty$ perturbations, we might have to involve higher-order information, which is usually very computationally expensive. In practice, those $l_\infty$ adversarially trained models can also be further attacked by perturbing few pixels with large $l_\infty$ perturbations, which still yields small $l_1$ or $l_2$ distance \cite{sharma2018attacking}. However, such a change sometimes even leads to misclassification by human. Moreover, it is unfair to compare an $l_1$ or $l_2$ attack with an $l_\infty$ attack (PGD) on $l_\infty$ adversarially trained models.

\section{Conclusion}
We generalize PGD on the space of data distributions, by learning an adversarial data distribution that maximally increases the generalization risk of a model. To solve the adversarial-distribution problem, we define a new energy functional to better reflex the discriminative data manifold in the WGF framework. When adopting particle approximation, adversarial samples can be generated by solving the corresponding WGF problems, leading to an algorithm closely related to the standard PGD method. Extensive evaluations show that our distributionally-adversarial attack outperforms PGD and Momentum PGD, achieving state-of-the-art attack results on the adversarially trained defense and provable defense models that demonstrated notable robustness against first-order $l_\infty$ attacks.

\bibliographystyle{aaai}
\bibliography{aaai}

\end{document}